%% file: main.tex
\documentclass[10pt,journal,compsoc]{IEEEtran}
\usepackage{hyperref}       %
\usepackage{url}            %
\usepackage{booktabs}       %
\usepackage{amsfonts}       %
\usepackage{graphicx}
\usepackage{amssymb, amsmath}
\usepackage{subfigure}
\usepackage{tablefootnote}
\usepackage{pdfpages}
\newcommand{\xhdr}[1]{\vspace{6pt}\noindent{\textbf{#1}}}

  \usepackage{cite}
\ifCLASSINFOpdf
\else
\fi
\begin{document}
	\title{Reinforcing Generated Images via Meta-learning for One-Shot Fine-Grained Visual Recognition}

	\author{Satoshi~Tsutsui,~%
		Yanwei~Fu,~%
		~David~Crandall,~\IEEEmembership{Member,~IEEE}%
		
		\IEEEcompsocitemizethanks{
		\IEEEcompsocthanksitem S. Tsutsui is with National University of Singapore.  E-mail: satoshi@nus.edu.sg

			\IEEEcompsocthanksitem Y. Fu is with School of Data Science, and
			 Shanghai Key Lab of Intelligent Information Processing, Fudan University, China. He is also with Fudan ISTBI—ZJNU Algorithm Centre for Brain-inspired Intelligence, Zhejiang Normal University, Jinhua, China.  E-mail: yanweifu@fudan.edu.cn. 
			 
		\IEEEcompsocthanksitem  D. Crandall is with Indiana University, Bloomington, USA.  E-mail: djcran@indiana.edu 
		
			 }
		
		\thanks{\scriptsize{Manuscript received 1 May 2021; revised 20 Nov. 2021; accepted 26 Mar. 2022. }}}

	\IEEEtitleabstractindextext{%
		\begin{abstract}
			One-shot fine-grained visual recognition often
			suffers from the problem of having few training examples for new fine-grained
			classes. To alleviate this problem, off-the-shelf image generation techniques based on Generative Adversarial Networks (GANs) can potentially create additional training images. However, these GAN-generated images are often not helpful for actually improving the accuracy of one-shot fine-grained
			recognition.  In this paper, we propose a meta-learning framework to
			combine generated images with original images, so that the resulting ``hybrid'' training images
			improve one-shot learning. Specifically, the generic image
			generator is updated by a few training instances of novel classes, and a
			Meta Image Reinforcing Network (MetaIRNet) is proposed to conduct
			one-shot fine-grained recognition as well as image reinforcement. Our experiments
			demonstrate consistent improvement over baselines on one-shot
			fine-grained image classification benchmarks. Furthermore, our analysis shows that the reinforced images have more diversity compared to the original and GAN-generated images.  
		\end{abstract}

		}

	\maketitle

	\IEEEdisplaynontitleabstractindextext
	\IEEEpeerreviewmaketitle

	\section{Introduction}
    The availability of vast labeled datasets has been crucial for the recent success of deep learning. However, there will always be learning tasks for which labeled data is sparse.  Fine-grained visual recognition is one typical example: when images are to be classified into many very specific categories (such as species of birds), it may be difficult to obtain training examples for rare classes, and producing the ground truth labels may require significant expertise (e.g., from ornithologists). One-shot learning is thus very desirable for fine-grained visual recognition.       
    
    A recent approach to address data scarcity is meta-learning~\cite{yuxiong2016eccv,santoro2016meta,finn2017model,img_deform_2019}, which trains a parameterized function called a meta-learner that maps labeled training sets to classifiers. The meta-learner is trained by sampling small training and test sets from a large dataset of a base class. Such a meta-learned model can be adapted to recognize novel categories with a single training instance per class.  Another way to address data scarcity is to synthesize additional training examples, for example by using Generative Adversarial Networks (GANs)~\cite{goodfellow2014generative,biggan}.  However, classifiers trained from GAN-generated images are typically inferior to those trained with real images, possibly because the distribution of generated images is biased towards frequent patterns ({modes}) of the original image distribution~\cite{shmelkov2018good}. This is especially true in one-shot {fine-grained} recognition where a tiny difference (e.g., beak of a bird) can make a large difference in class.

	\begin{figure}
		\centering
		\includegraphics[width=0.95\columnwidth]{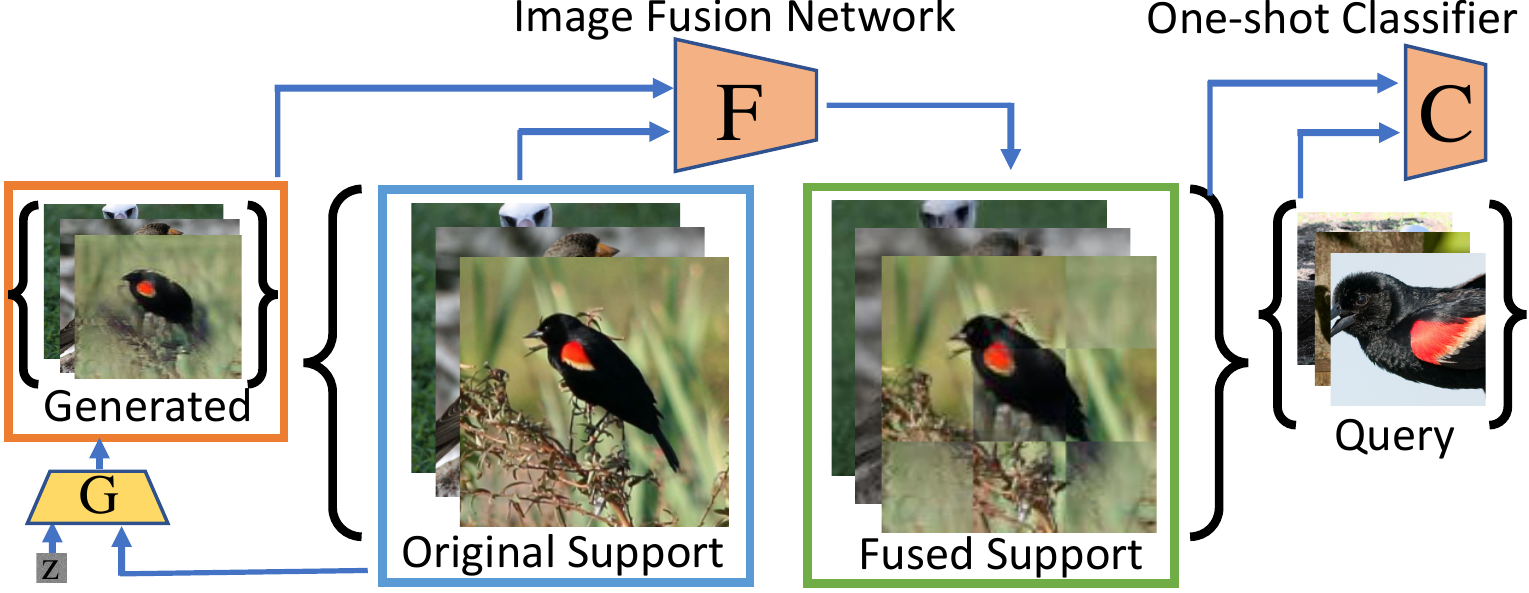}
		\caption{\textbf{Our Meta Image Reinforcing Network (MetaIRNet) } consists of an image fusion network and a one-shot classifier. The image fusion network reinforces generated
			images to  make them more beneficial for the one-shot classifier by diversifying the images (Figure~\ref{fig:analysis-dist}), while the
			one-shot classifier learns representations that are suitable to
			classify unseen examples with few examples. Both networks
			are trained  end-to-end, so the loss back-propagates from
			classifier to the fusion network.}\label{fig:method}
	\end{figure}

	\begin{figure}
	\centering
	\includegraphics[width=0.96\columnwidth]{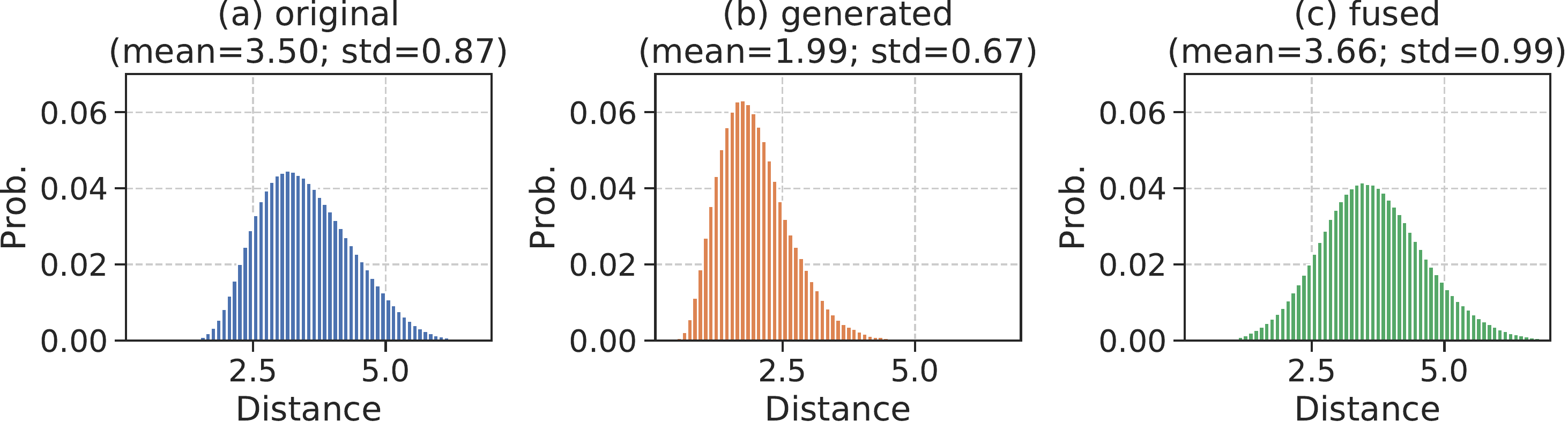}
	\caption{Distribution of pairwise distances for (a) original set, (b) generated set, and (c) fused set. Our fused images have greater diversity, while generated images are not as diverse as the originals. }\label{fig:analysis-dist}
\end{figure}

     In this paper, we develop an approach to apply   off-the-shelf generative models to synthesize training data in a  way that improves one-shot fine-grained classifiers (Fig.~\ref{fig:method}). We begin by conducting a pilot study in which we investigate using a  generator pre-trained on ImageNet in a one-shot scenario. We show that the generated images can indeed improve the performance of a one-shot classifier when used with a manually designed rule to combine the generated images with the originals using the weights of a $3\times3$ block matrix (like Fig. \ref{fig:fintune-gan-samples} (g)). These preliminary results lead us to consider optimizing these block matrices in a data-driven manner. Thus, we propose a meta-learning approach to learn these block matrices to reinforce the generated images effectively for few-shot classification. 
     
     Our approach has two steps. First, an off-the-shelf generator trained  from ImageNet is updated towards the domain of novel classes by using  only a single image (Sec.~\ref{sec:method-gen}). Second,  since previous  work and our pilot study (Sec.~\ref{sec:pilot-study}) suggest that  simply adding synthesized images to the training data may not improve  one-shot learning, the synthesized images are ``mixed'' with the  original images in order to bridge the domain gap between the two  (Sec.~\ref{sec:method-mix}). The effective mixing strategy is learned  by a meta-learner, which essentially  boosts the performance of fine-grained categorization with  a single training instance per class. We experimentally validate that our  approach can achieve improved performance over  baselines on  fine-grained classification datasets in one-shot situations  (Sec.~\ref{sec:experiment}). Moreover, we empirically analyze the mixed images and investigate how our learned mixing strategy reinforces the original images (Sec.~\ref{sec:analysis}). As highlighted in Figure~\ref{fig:analysis-dist}, we show that while the GAN-generated images lack  diversity compared to the original, our mixed images effectively introduce additional diversity.

    \textbf{Contributions} of this paper are that we: 1) Introduce a method to transfer a pre-trained generator with a single image; 2) Propose a meta-learning method to learn to complement real images with synthetic images in a way that benefits one-shot classifiers; 3) Demonstrate that these methods improve one-shot classification accuracy on fine-grained visual recognition benchmarks;  and 4)  Analyze our resulting mixed images and empirically show that our method can help diversify the dataset.  
    A preliminary version of this paper appeared in NeurIPS~\cite{metasatoshi19}. %
	\section{Related Work}\label{sec:related}
	Our paper relates to three main lines of work: GAN-synthesized images for training, few-shot meta-learning, and data augmentation to diversify training examples.
	
	\subsection{Image Generation by GANs} 
	Learning to generate realistic
	images is challenging  because it is difficult to
	define a loss function that accurately measures perceptual photo realism.
	Generative Adversarial
	Networks (GANs)~\cite{goodfellow2014generative} address this issue by
	learning not only a generator but also a loss function --- the
	discriminator --- that helps the generator to synthesize images
	indistinguishable from real ones.  
	This
	adversarial learning is intuitive but is often 
	unstable in practice~\cite{gulrajani2017improved}. Recent progress
	includes better CNN
	architectures~\cite{radford2015unsupervised,biggan}, training
	stabilization~\cite{arjovsky2017wasserstein,gulrajani2017improved,miyato2018spectral,wang2020stabilizing},
	and exciting applications (e.g.~\cite{ojha2021few-shot-gan,park2020swapping}).  BigGAN~\cite{biggan} trained on ImageNet has shown
	visually-impressive generated images with stable performance on
	generic image generation tasks.  Several
	studies~\cite{noguchi2019image,wang2018transferring} have explored
	generating images from few examples, but their focus has not been on
	one shot classification.  Several
	papers~\cite{de2017modulating,dumoulin2017learned,noguchi2019image}
	also use the idea of adjusting batch normalization layers, which helped
	to inspire our work. Finally, work has investigated using GANs to
	help image
	classification~\cite{shmelkov2018good,shrivastava2017learning,antoniou2018augmenting,zhang2018metagan,gao2018low};
	ours differs in that we apply an off-the-shelf generator
	pre-trained from a large and generic dataset.
	
	\subsection {Few-shot Meta-learning} 
	Few shot
	classification~\cite{chen2019closer} with meta-learning has received much attention after the introduction of MetaDataset~\cite{Triantafillou2020MetaDatasetAD}.
	The task is to train a
	classifier with only a few examples per class.  Unlike the typical
	classification setup, the classes in the
	training and test sets have no overlap, and the
	model is trained and evaluated by sampling many few-shot tasks (or
	episodes). For example, when training a dog breed classifier, an
	episode might train to recognize five dog species with only a single training
	image per class --- a 5-way-1-shot setting.  A meta-learning method
	trains a meta-model by sampling many episodes from training classes
	and is evaluated by sampling many episodes from other unseen
	classes. With this episodic training, we can choose several possible
	approaches to ``learn to learn.'' For example, ``learning to compare''
	methods learn a metric space
	(e.g.,~\cite{vinyals2016matching,snell2017prototypical,sung2018learning,bateni2020improved}),
	while other approaches learn to fine-tune (e.g.,~\cite{finn2017model,rusu2018meta,finn2018probabilistic,ravi2017optimization})
	or learn to augment data (e.g.,~\cite{wang2018low,hariharan2017low,chen2019imageblock_aaai,gao2018low, schwartz2018delta}).  Our approach
	also explores data augmentation by mixing the original images with
	synthesized images produced by a fine-tuned generator, but we find
	that the naive approach of simply adding GAN-generated images to the
	training dataset does not improve performance.  However, by carefully
	combining generated images with the original images, we find that
	we can  synthesize examples that  do increase
	the performance. Thus we employ meta-learning to learn the proper combination strategy.
	
	\subsection{Data Augmentation}
	Data augmentation is often an integral part of 
	training deep CNNs; in fact, AlexNet~\cite{krizhevsky2012imagenet} 
	describes data augmentation as one of ``the two primary ways in 
	which we combat overfitting.'' Since then, data augmentation strategies
	have been explored~\cite{zhong2020random}, but 
	they are manually designed and thus not scalable for many 
	domain-specific tasks. Recent work uses automated 
	approaches to search for the optimal augmentation 
	policy using reinforcement learning~\cite{cubuk2019autoaugment} or 
	by directly optimizing the augmentation policy by making it differentiable~\cite{li2020differentiable}. 
	Moreover, some researchers perform empirical analysis on data 
	augmentation as a distributional shift~\cite{gontijo-lopes2021tradeoffs}, or develop
	a theoretical framework of data augmentation as a Markov process~\cite{dao2019kernel}. 
	Our work is most closely related to augmentation based on mixing images.
For example, Mixup~\cite{zhang2018mixup} linearly mixes two random images with a random weight. Manifold Mixup performs a similar operation to the CNN representation of the images~\cite{manifoldmixup19}. CutMix~\cite{yun2019cutmix} overlays
randomly-cropped images onto other images at random locations. While these methods randomly mix images while ignoring the content of them, our method learns to adjust the  mixing technique for given images via meta-learning that optimizes the parameterized mixing strategy to help one-shot learning.

	\section{Pilot Study}
	\label{sec:pilot-study} 
	
	To explain how we arrived at our approach, we describe the initial
	experimentation which motivated our methods.
	
	\newcommand{\pilotfigw}{0.1}
	\begin{figure}
		\centering
		\vspace{-1mm}
		\begin{subfigure}[]{%
				\includegraphics[clip, width= \pilotfigw\textwidth]{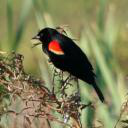}}
		\end{subfigure}
		\vspace{-1mm}
		\begin{subfigure}[]{%
				\includegraphics[clip, width= \pilotfigw\textwidth]{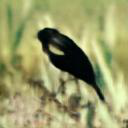}}
		\end{subfigure}
		\begin{subfigure}[]{%
				\includegraphics[clip, width= \pilotfigw\textwidth]{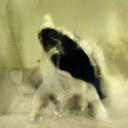}}
		\end{subfigure}
		\begin{subfigure}[]{%
				\includegraphics[clip, width= \pilotfigw\textwidth]{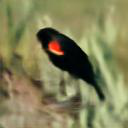}}
		\end{subfigure}
		\begin{subfigure}[]{%
				\includegraphics[clip, width= \pilotfigw\textwidth]{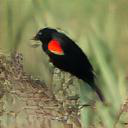}}
		\end{subfigure}
		\begin{subfigure}[ ]{%
				\includegraphics[clip, width= \pilotfigw\textwidth]{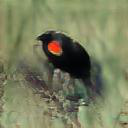}}
		\end{subfigure}
		\begin{subfigure}[]{%
				\includegraphics[clip, width= \pilotfigw\textwidth]{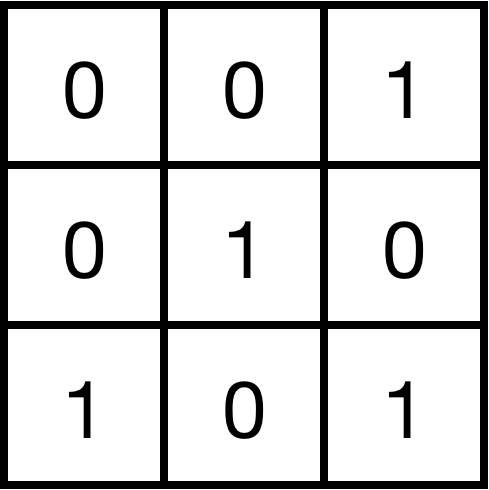}}
		\end{subfigure}
		\begin{subfigure}[ ]{%
				\includegraphics[clip, width= \pilotfigw\textwidth]{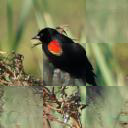}}
		\end{subfigure}
		\caption{Samples described in Sec. \ref{sec:pilot-study}. (a)
			Original image. (b) Result of tuning noise only. (c) Result of tuning the whole
			network. (d) Result of tuning batch norm only. (e) Result of tuning batch norm with
			perceptual loss. (f) Result of slightly disturbing noise from (e). (g) a
			$3 \times 3$ block weight matrix $w$. (g) Result of mixing (a) and (f) as
			$w \times$(f) + $(1 - w )\times$(a). 
		}\label{fig:fintune-gan-samples}
	\end{figure}
	
	\begin{table}
		\caption{CUB  5-way-1-shot classification accuracy (\%)
			using ImageNet features. Simply adding generated images to the training
			set does not help, but adding hybrid images, as
			in Fig. \ref{fig:fintune-gan-samples} (h), can.}
		\label{tbl:pilot-study}
		\centering
		{\scriptsize{
		\begin{tabular}{l@{\,\,\,\,\,\,\,\,\,\,}cccc}
			\toprule
			       &  Nearest  & Logistic  &  Softmax  \\
                        Training Data      &  Neighbor  & Regression &  Regression  \\
			\midrule
            Original &  $70.72 \pm 0.51$ &  $75.39 \pm 0.47$ &  $74.61 \pm 0.48$ \\
            Original + Generated  &  $70.84 \pm 0.51$ &  $74.08 \pm 0.48$ &  $73.55 \pm 0.48$ \\
            Original + Mixed   &  $71.50 \pm 0.50$ &  $76.07 \pm 0.47$ &  $75.40 \pm 0.47$ \\
			\bottomrule
		\end{tabular}
		}}
	\end{table}

	\subsection{How to transfer knowledge from
		pre-trained GANs?}  We aim to quickly generate training images for
	few-shot classification. Performing adversarial learning
	(\emph{i.e}., training a generator and discriminator initialized with
	pre-trained weights) is not practical when we only have one or two
	examples per class. Instead, we want to develop a method that does not
	depend on the number of images at all; in fact, we consider the
	extreme case where only a single image is available, and want to
	generate variants of the image using a pre-trained GAN. We tried fixing
	the generator weights and optimizing the noise so that it generates
	the target image, under the assumption that sightly modifying the
	optimized noise would produce a variant of the original. However,
	naively implementing this idea with BigGAN did not reconstruct the
	image well, as shown in the example in
	Fig.~\ref{fig:fintune-gan-samples}(b). We then tried also fine-tuning the generator
	weights, but this produced even worse images stuck in a local
	minimum, as shown in Fig~ \ref{fig:fintune-gan-samples}(c).  
	
	We speculate that the best approach may be somewhere in between the two
	extremes of tuning noise only and tuning both noise and
	weights. Inspired by previous
	work~\cite{de2017modulating,dumoulin2017learned,noguchi2019image}, we
	propose to fine-tune only scale and shift parameters in the batch
	normalization layers. This strategy produces better images, as shown in
	Fig.~\ref{fig:fintune-gan-samples}(d). Finally, again inspired by
	previous work~\cite{noguchi2019image}, we not only minimize the
	pixel-level distance but also the distance of a pre-trained CNN
	representation (perceptual loss~\cite{johnson2016perceptual}),
	yielding slightly improved results (Fig.
	\ref{fig:fintune-gan-samples}(e)). We can  generate slightly
	different versions by adding random perturbations to the tuned noise
	(e.g., the ``fattened'' version of the same bird in
	Fig.~\ref{fig:fintune-gan-samples}(f)).  The entire training process
	requires fewer than 500 iterations and takes less than 20 seconds on an
	NVidia Titan Xp GPU. We explain the generation strategy that we developed based on this pilot study in
	Sec.~\ref{sec:method}.
	
	\subsection{Do generated images help few-shot learning?}
	Our goal is not to generate images, but to augment the training data
	for few shot learning. A naive way to do this is to apply the above generation
	technique for each training image, in order to double the training set. We
	tested this idea on a validation set (split the same as \cite{chen2019closer}) from
	the Caltech-UCSD bird dataset~\cite{WahCUB_200_2011} and computed mean accuracy and 95\% confidence intervals on 2000 episodes of
	5-way-1-shot classification. We used pre-trained ImageNet features from
	ResNet18~\cite{he2016deep} with nearest neighbor, one-vs-all logistic
	regression, and softmax regression (or multi-class logistic regression).
	As shown in Table~\ref{tbl:pilot-study}, the accuracy actually drops for two of the three classifiers when we double
	the size of our training set by generating synthetic training images, suggesting
	that the
	generated images are harmful for training classifiers. 

	\subsection{How to synthesize images for
		few-shot learning?} Given that the synthetic images
	\textit{appear} meaningful to humans, we conjecture that they can
	benefit few shot classification when properly mixed with originals to create hybrid images. To empirically test this hypothesis, we devised a random $3\times3$
	grid to combine the images, which is inspired by $3\times3$ visual jigsaw pretraining~\cite{noroozi2016unsupervised}. As shown in
	Fig. \ref{fig:fintune-gan-samples}(h), images (a) and (f) were combined by
	taking a linear combination within each cell of the grid shown in (g).
	Finally, we added mixed images like (h) into the training data, and
	discovered that this produced a modest increase in accuracy (last row
	of Table \ref{tbl:pilot-study}). While the increase is marginal, these
	mixing weights were binary and manually selected, and thus likely not
	optimal.  In Sec.~\ref{sec:method-mix}, we show how to learn this
	mixing strategy in an end-to-end manner using a meta-learning
	framework.

	\section{Method}
	\label{sec:method}
	The results of the pilot study in the last section suggested that 
	producing synthetic images could be useful for few-shot fine-grained
	recognition, but only if  done in a careful way. In this section,
	we use these findings to propose a novel technique that does this
	effectively (Fig.~\ref{fig:method}).
	We
	propose a GAN fine-tuning method that works with a single image
	(Sec.~\ref{sec:method-gen}), and a meta-learning method to not
	only learn to classify with few  examples, but also to learn to
	 reinforce the generated images
	(Sec.~\ref{sec:method-mix}).

	\subsection{FinetuneGAN: Fine-tuning Pre-trained Generator for Target Images}
	\label{sec:method-gen} 
	
	GANs typically have a generator $G$ and a discriminator $D$. Given an input
	signal $z\sim\mathcal{N}(0,1)$, a well-trained generator synthesizes
	an image $G(z)$. In our tasks, we adapt an off-the-shelf GAN generator
	$G$ that is pre-trained on the ImageNet-2012 dataset in order to generate
	more images in a target, data-scarce domain.  Note that we do not use the
	discriminator, since adversarial training with a few images is
	unstable and may lead to model collapse. Formally, we fine-tune $z$ and
	the generator $G$   such that $G$ generates an image $\mathbf{I}_{z}$
	from an input vector $z$ by minimizing the distance between $G(z)$ and
	$\mathbf{I}_{z}$, where the vector $z$ is randomly
	initialized. Inspired by previous
	work~\cite{noguchi2019image,arjovsky2017wasserstein,johnson2016perceptual},
	we minimize a loss function $\mathcal{L}_{G}$ with 
	$\mathcal{L}_{1}$ distance and perceptual loss $\mathcal{L}_{perc}$
	with earth mover regularization $\mathrm{\mathcal{L}}_{EM}$,
	\begin{align}
		\mathcal{L}_{G}\left(G,\mathbf{I}_{z},z\right)=\mathcal{L}_{1}\left(G(z),\mathbf{I}_{z}\right)+\lambda_{p}\mathcal{L}_{perc}\left(G(z),\mathbf{I}_{z}\right) \nonumber \\ +\lambda_{z}\mathrm{\mathcal{L}}_{EM}\left(z,r\right), \label{eq:loss-gen}
	\end{align}
	where $\lambda_{p}$ and $\lambda_{z}$
	are coefficients of each term. The first term is the L1 distance of $G(z)$ and $\mathbf{I}_{z}$ using pixels. The second term is basically L2 distance of $G(z)$ and $\mathbf{I}_{z}$ but using, instead of pixels, the intermediate feature maps from all convolution layers of ImageNet-trained VGG16~\cite{vgg}. The last term is the earth mover distance~\cite{arjovsky2017wasserstein} of 	$z$ and $r\sim\mathcal{N}(0,1)$ (random noise sampled from the normal distribution).
	
	Since only a few training images are available in the target domain, only
	scale and shift parameters of the batch normalization of $G$ are updated
	in practice. Specifically, only the $\gamma$ and $\beta$
	of each batch normalization layer are updated in each layer,
	\begin{align}
		\gamma \left( \frac{x-\mathrm{\mathbb{E}}(x)}{\sqrt{\mathrm{Var}(x)+\epsilon}} \right)+\beta,\label{eq:scale=000026gamma}
	\end{align}
	\noindent where $x$ is the input feature from the previous layer, and $\mathbb{E}$
	and $\mathrm{Var}$ indicate the mean and variance functions, respectively. Intuitively
	and in principle, updating $\gamma$ and $\beta$ only is equivalent
	to adjusting the activation of each neuron in a layer.  Once updated,
	the $G(z)$ would be synthesized to reconstruct the image $\mathbf{I}_{z}$.
	Empirically, a small random perturbation $\epsilon$ is added to
	$z$ as $G\left(z+\epsilon\right)$.
	Examples of $\mathbf{I}_{z}$, $G(z)$ and $G\left(z+\epsilon\right)$
	are illustrated in in Fig. \ref{fig:fintune-gan-samples} (a), (e),
	and (f), respectively.
	
	\subsection{Meta-Reinforced Synthetic Data  \label{sec:method-mix}}

	\subsubsection{One-shot Learning Defined} One-shot classification is a
	meta-learning problem that divides a dataset into two sets:
	meta-training (or base) set and meta-testing (or novel) set. The
	classes in the base and novel sets are disjoint. In other
	words, 
	\begin{align}
		\mathcal{D}_{base}=\left\{
		\left(\mathbf{I}_{i},y_{i}\right), y_{i}\in\mathcal{C}_{base}\right\}, \\
		\mathcal{D}_{novel}=\left\{ \left(\mathbf{I}_{i},y_{i}\right),
		y_{i}\in\mathcal{C}_{novel}\right\},
	\end{align}
	where $\mathcal{C}_{base}\cup
	\mathcal{C}_{novel}=\emptyset$.  
	
	The task is to train a classifier on
	$\mathcal{D}_{base}$ that can quickly generalize to unseen classes in
	$\mathcal{C}_{novel}$ with one or few examples.  To do this, a
	meta-learning algorithm performs meta-training by sampling many
	one-shot tasks from $\mathcal{D}_{base}$, and is evaluated by sampling
	many similar tasks from $\mathcal{D}_{novel}$. Each sampled task
	(called an episode) is an $n$-way-$m$-shot classification problem with
	$q$ queries, meaning that we sample $n$ classes with $m$ training and
	$q$ test examples for each class.  In other words, an episode has a
	support (or training) set $S$ and a query (or test) set $Q$, where
	$|S|=n \times m$ and $|Q|=n\times q$. One-shot learning means $m=1$. The
	notation $S_c$ means the support examples only belong to the class
	$c$, so $|S_c| = m$.

	\subsubsection{Meta Image Reinforcing Network (MetaIRNet).} We
	propose a Meta Image Reinforcing Network (MetaIRNet), which not only
	learns a few-shot classifier, but also learns to reinforce generated
	images by combining real  and 
	generated images. MetaIRNet is composed of two modules: an image fusion
	network $F$, and a one-shot classifier $C$.

	\xhdr{Image Fusion Network}
	$F$ combines a real image
	$\mathbf{I}$ and a corresponding generated image
	$\mathbf{I}_{g}$ into a new image $\mathbf{I}_{syn}=F\left(\mathbf{I},\mathbf{I}_{g}\right)$, which will be added into the support set. Note that for each real image (regardless of whether it is a positive or negative example) in the support set, we use an image generated by FinetuneGAN for mixing. While there could be many possible ways to mix the two images (i.e., the design decision of $F$), 
	we were inspired by $3\times3$ visual jigsaw pretraining~\cite{noroozi2016unsupervised} and
	its data augmentation applications~\cite{img_deform_2019}. Thus, as shown in
	Figure~\ref{fig:fintune-gan-samples}(g), we divide the images into a
	$3\times3$ grid and linearly combine the cells with the weights
	$\mathbf{w}$ produced by a CNN conditioned on the two images,
	\begin{equation}
		\mathbf{I}_{syn}=\mathbf{w} \odot \mathbf{I}+\left(1-\mathbf{w}\right) \odot \mathbf{I}_{g},
	\end{equation}
	where $\odot$ is element-wise multiplication, and $\mathbf{w}$ is
	resized to the image size keeping the block structure. The CNN that produces $\mathbf{w}$ extracts the feature vectors of $\mathbf{I}$ and
	$\mathbf{I}_g$, concatenates them, and uses a fully-connected
	layer to produce a weight corresponding to each of the nine cells in the $3
	\times 3$ grid.  Finally, for each real image
	$\mathbf{I}^{i}$, we generate $n_{aug}$ synthetic images, and assign the same class label $y_{i}$ to 
	each synthesized image $\mathbf{I}_{syn}^{i,j}$ to obtain an augmented
	support set,
	\begin{equation}
		\tilde{S}=\left\{ \left( \mathbf{I}^{i}_{},y^{i}_{}\right),\left\{ \left(\mathbf{I}_{syn}^{i,j},y_{}^{i}\right)\right\}_{j=1}^{n_{aug}}\right\}_{i=1}^{n\times m} .
	\end{equation}
	
	\xhdr{One-Shot Classifier}
	$C$ maps an input image $\mathbf{I}$ into
	feature maps $C\left(\mathbf{I}\right)$, and performs
	one-shot classification. Although any one-shot classifier can be
	used, we choose the non-parametric prototype classifier
	of Snell \textit{et al.}~\cite{snell2017prototypical} due to its superior performance and
	simplicity. During each episode, given the sampled $S$ and $Q$,
	the image fusion network produces an augmented support set
	$\tilde{S}$. This classifier computes the prototype vector $\mathbf{p}_c$ for each class $c$ in $\tilde{S}$ as an average feature
	vector, 
	\begin{equation}
		\mathbf{p}_c = \frac{1}{|\tilde{S_c}|}\sum_{(\mathbf{I}_{i},y_{i})\in\tilde{S_c}}C\left(\mathbf{I}_{i}\right). 
	\end{equation}
	
	For a query image $\mathbf{I}_{i}\in Q$,
	the probability of belonging to a class $c$ is estimated as, 
	\begin{equation}
		P\left(y_{i}=c\text{\ensuremath{\mid}}\mathbf{I}_{i}\right)=
		\frac{\mathrm{exp}\left(-\left\Vert  C\left(\mathbf{I}_{i}\right)-\mathbf{p}_c\right\Vert \right)}{\sum_{k=1}^{n}\mathrm{exp}\left(-\left\Vert C\left(\mathbf{I}_{i}\right)-\mathbf{p}_k\right\Vert \right)},\label{eq:prototypical_classifier}
	\end{equation}
	\noindent where $\parallel\cdot\parallel$ is the Euclidean
	distance. Then, for a query image, the class with the highest
	probability becomes the final prediction of the one-shot classifier.
	
	\xhdr{Training.} In the meta-training phase, we jointly train  $F$ and $C$  end-to-end, minimizing a cross-entropy loss,
	\begin{equation}
		\min_{\mathbf{\theta}_F,\mathbf{\theta}_C}
		\frac{1}{|Q|}\sum_{\left(\mathbf{I}_{i},y_{i}\right)\in
			Q}-\mathrm{log}P\left(y_{i}\mid\mathbf{I}_{i}\right),\label{eq:loss_function}
	\end{equation} 
	where $\mathbf{\theta}_F$ and $\mathbf{\theta}_C$ are the learnable parameters of $F$ and $C$.

\input{table1}

	\section{Experiments}\label{sec:experiment}
	To investigate the effectiveness of our approach, we perform 1-shot-5-way
	classification  following the meta-learning experimental
	setup described in Sec. \ref{sec:method-mix}.  We perform 1000 episodes
	in meta-testing, with 16 query images per class per episode, and
	report average classification accuracy and 95\% confidence
	intervals. We use the fine-grained classification dataset of Caltech UCSD
	Birds (CUB)~\cite{WahCUB_200_2011} for our main experiments,
	and another fine-grained dataset, North American Birds
	(NAB)~\cite{van2015building}, for secondary experiments. CUB has 11,788
	images with 200 classes, and NAB has 48,527 images with 555 classes.

	\subsection{Implementation Details}
     While our fine-tuning method introduced in Sec.~\ref{sec:method-gen} can generate images for each step in meta-training and meta-testing, it takes around 20 seconds per image, so we apply the generation method ahead of time to make our experiments more efficient. This means that the generator is trained independently. We use a BigGAN pre-trained on ImageNet, using the publicly-available weights. We set $\lambda_p = 0.1$ and $\lambda_z = 0.1$, and perform 500 gradient descent updates with the Adam~\cite{kingma2014adam} optimizer with learning rate $0.01$ for $z$ and $0.0005$ for the fully connected layers, to produce scale and shift parameters of the batch normalization layers.  We manually chose these hyper-parameters by trying random values from 0.1 to 0.0001 and visually checking the quality of a few generated images. We only train once for each image, generate 10 random images by perturbing $z$, and randomly use one of them for each episode ($n_{aug} = 1$). For image classification, we use ResNet18~\cite{he2016deep} pre-trained on ImageNet for the two CNNs in $F$ and one in $C$. Note that we do not share weights among the three CNNs, which means that our model has three ResNets inside. We train $F$ and $C$ with Adam with a default learning rate of $0.001$. We select the best model based on the   validation accuracy, and then compute the final accuracy on the test   set. We use the same train/val/test split used in previous   studies~\cite{chen2019closer,metasatoshi19} for CUB and NAB, respectively. Further implementation details are available as supplemental source code.\footnote{http://vision.soic.indiana.edu/metairnet/}
	
	\subsection{Baselines}
	\xhdr{Non-meta learning classifiers.} We directly train the same ImageNet pre-trained CNN used
	in $F$ to classify images in $\mathcal{D}_{base}$, and use it as a
	feature extractor for $\mathcal{D}_{novel}$. We then use the following off-the-shelf
	classifiers:
		(1) \textbf{Nearest Neighbor};
		(2) \textbf{Logistic Regression} (one-vs-all classifier);
		(3) \textbf{Softmax Regression} (also called multi-class logistic regression).
	
	\xhdr{Meta-learning classifiers.} We
	try the meta-learning method of prototypical network
	(\textbf{ProtoNet}~\cite{snell2017prototypical}).  ProtoNet computes
	an average prototype vector for each class and performs nearest
	neighbor with the prototypes. We note that our MetaIRNet adapts
	ProtoNet as a choice of $F$ so this is an ablative version of our
	model (MetaIRNet without the image fusion module).
		
	\xhdr{Data augmentation.}  We compare against simply using the generated images as data augmentation, as well as applying typical data augmentations. Moreover, because our MetaIRNet uses meta-learning to find the best way to mix the original and GAN-generated images, we compare against several alternative ways of mixing them:		
    (1) \textbf{Flip} horizontally flips the images;
	(2) \textbf{Gaussian} adds Gaussian noise with standard deviation 0.01 into the CNN features;
	(3) \textbf{FinetuneGAN}~(introduced in Sec. \ref{sec:method-gen}) generates augmented images by fine-tuning the ImageNet-pretrained BigGAN with each support set;
	(4) \textbf{Mixup}~\cite{zhang2018mixup} mixes two images with a randomly sampled weight;
	(5) \textbf{Manifold Mixup}~\cite{manifoldmixup19} does mixup in the CNN representation of images; and
	(6) \textbf{CutMix}~\cite{yun2019cutmix} mixes the two images with  randomly sampled locations. 
    We do these augmentations in the meta-testing stage to increase the support set. For fair comparison, we use ProtoNet as the base classifier of all these baselines.
    
    \xhdr{Mix with other images.} To evaluate the utility of our
    generated images, we use our meta-learning technique (MetaIRNet) to mix with
    images that are \textit{not} from our FinetuneGAN: (1)
    \textbf{FreezeDGAN}~\cite{mo2020freeze} fine-tunes GANs by
    performing adversarial training using a stabilization technique of
    freezing the discriminator; and (2) \textbf{Jitter} produces
    data-augmented images by randomly jittering the original images.

	\subsection{Results}
	As shown in Table~\ref{tbl:results-main}, our MetaIRNet is superior to
	all baselines including the meta-learning classifier of ProtoNet
	(84.13\% vs. 81.73\%) on the CUB dataset. It is notable that while ProtoNet
	has worse accuracy when simply using the generated images as
	data augmentation, our method shows an accuracy increase from ProtoNet,
	which is equivalent to MetaIRNet without the image fusion module. This
	indicates that our image fusion module can effectively complement the
	original images while removing harmful elements from generated ones.
	Interestingly, horizontal flip augmentation yields nearly a 1\% accuracy
	increase for ProtoNet. Because flipping cannot be learned
	directly by our method, we conjectured that our method could also
	benefit from it. The final row of the table shows an additional experiment
	with our MetaIRNet
	combined with random flip augmentation, showing an additional accuracy increase  from 84.13\% to 84.80\%. This suggests that our method
	provides an improvement that is orthogonal to flip
	augmentation. 
	
	Lastly, while most of our experiments focus on the 1-shot cases, we also tested 5-shot  and obtained an accuracy of $93.09 \pm 0.30$\%, which is higher than baselines. More details are in Supplementary Material.
	
\begin{figure}[t!]
	\centering
	\begin{minipage}{0.48\columnwidth}
		\centering
		\includegraphics[clip, width=0.85\textwidth] {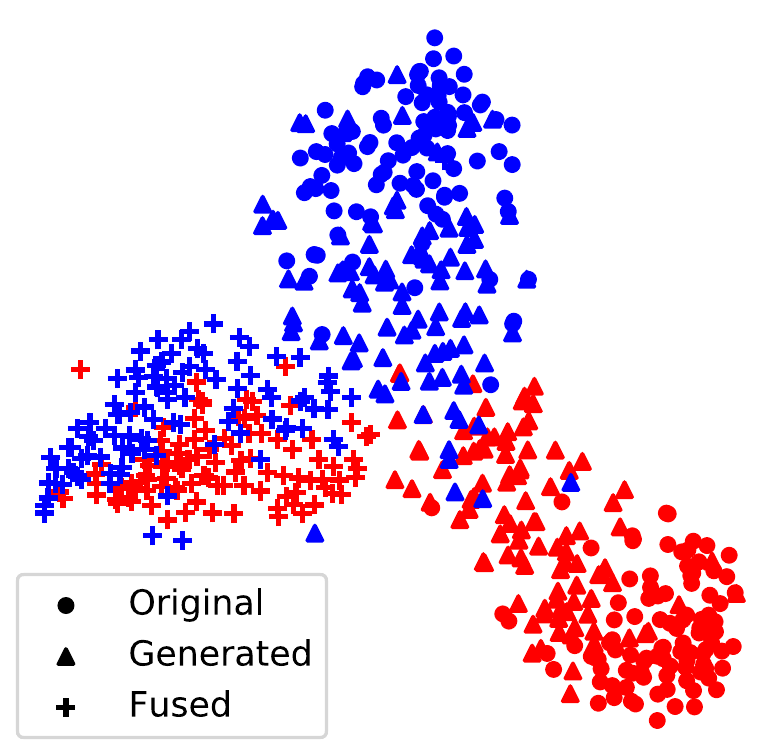}
		\caption{ t-SNE plot of two random classes. We use colors (red and blue) to represent the classes, and use different markers for the three types of images -- real
			images (\textbullet), generated images ($\blacktriangle$), and
			augmented fused images (\textbf{+}) }\label{fig:tsne}
	\end{minipage}\hfill
	\begin{minipage}{0.48\columnwidth}
		\centering
		\includegraphics[clip, width=\textwidth ]{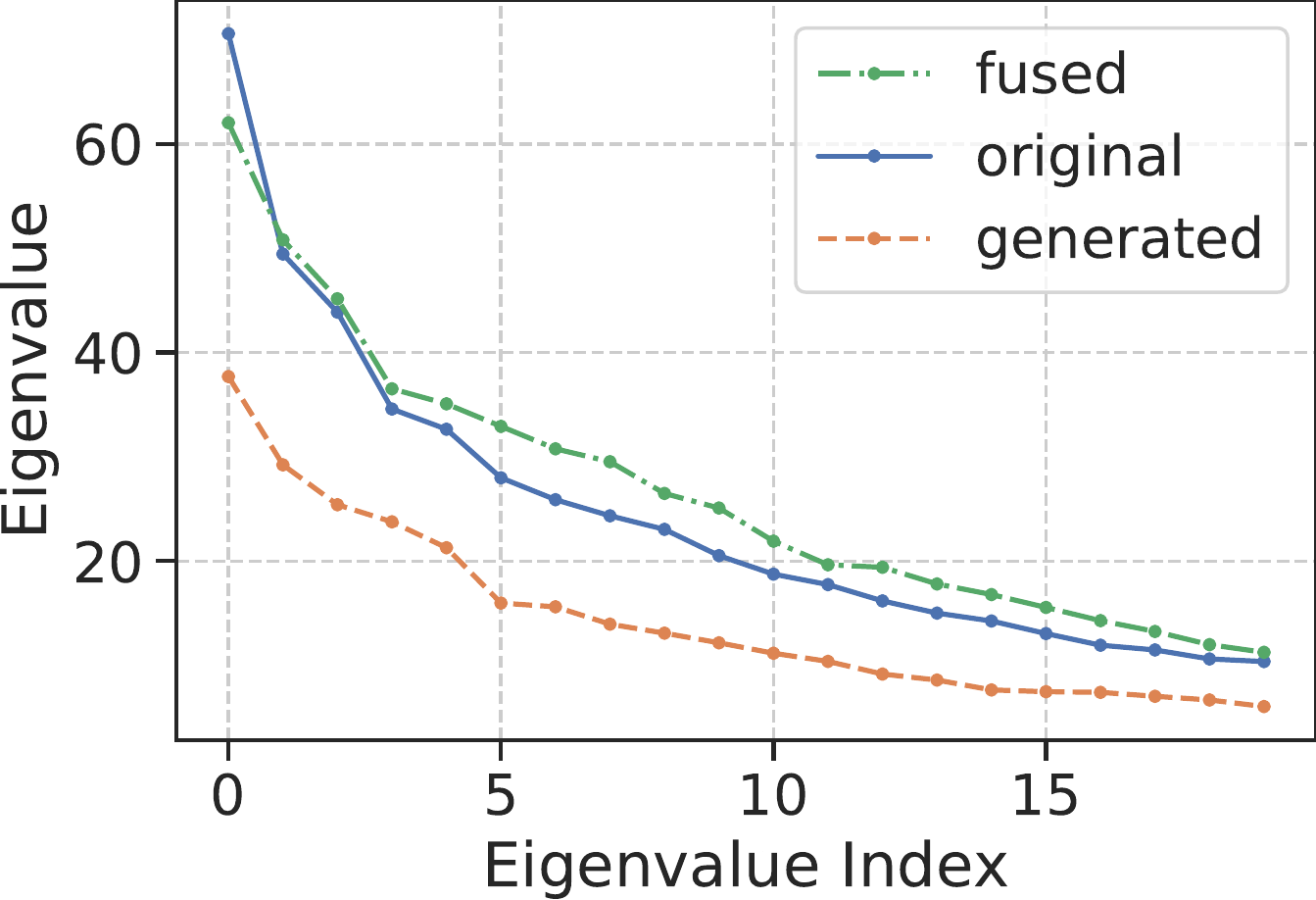} 
		\caption{
			The sorted eigenvalues of principal component analysis (PCA) for each set. The higher eigenvalues mean that the manifold in the the feature space is wider, which suggests greater diversity. Our fused images are at least as diverse as original.  
		}\label{fig:result-eigenvalue}
	\end{minipage}
\end{figure}

	\xhdr{Case Studies.}  
	We show some sample visualizations in
	Fig.~\ref{fig:result-samples}. We observe that image generation often
	works well but sometimes completely fails. An advantage of our
	technique is that even in these failure cases, our fused images often
	maintain some of the object's shape, even if the images themselves do
	not look realistic.  In order to investigate the quality of generated
	images in more detail, we randomly pick two classes,
	sample 100 images for each class, and show a t-SNE visualization of real
	images (\textbullet), generated images ($\blacktriangle$), and
	augmented fused images (\textbf{+}) in Fig.  \ref{fig:tsne}, with classes shown in red and blue. It is
	reasonable that the generated images are closer to the real ones,
	because our loss function in Equation (1) encourages this to be
	so. Interestingly, perhaps due to artifacts of $3 \times 3$ patches,
	the fused images are distinctive from the real and generated images,
	which extends the decision boundary.

	\newcommand{\imgwidtha}{0.17}
	\newcommand{\imgwidthb}{0.22}
	
	\begin{figure}
		\centering
		\begin{tabular}{@{}c@{\,\,\,}c@{\,\,\,}c@{\,\,\,}c@{}}
			Original & Generated & Fused & Weight    \\
			
			\includegraphics[width=\imgwidtha \linewidth ]
			{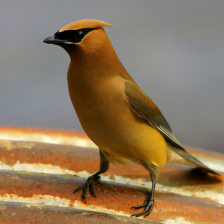}& 
			\includegraphics[width=\imgwidtha \linewidth ]
			{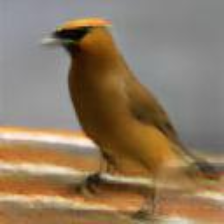}&
			\includegraphics[width=\imgwidtha \linewidth ]
			{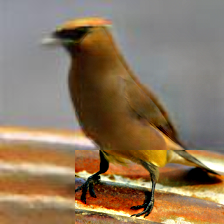}&
			\includegraphics[width=\imgwidthb \linewidth ]
			{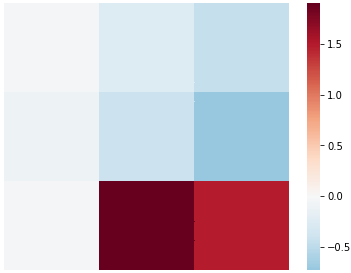}
			\\
			
			\includegraphics[width=\imgwidtha \linewidth ]
			{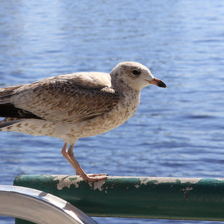}& 
			\includegraphics[width=\imgwidtha \linewidth ]
			{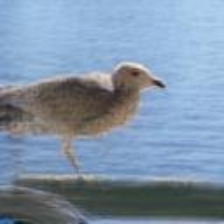}&
			\includegraphics[width=\imgwidtha \linewidth ]
			{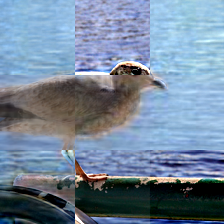}&
			\includegraphics[width=\imgwidthb \linewidth ]
			{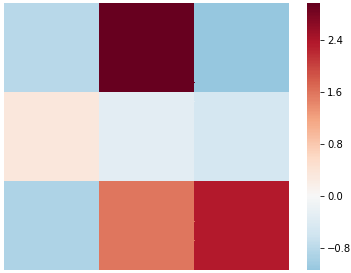}
			\\
			
			\includegraphics[width=\imgwidtha \linewidth ]
			{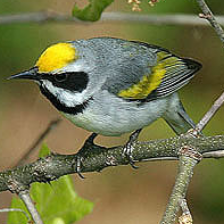}&
			\includegraphics[width=\imgwidtha \linewidth ]
			{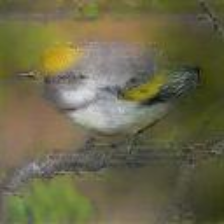}&
			\includegraphics[width=\imgwidtha \linewidth ]
			{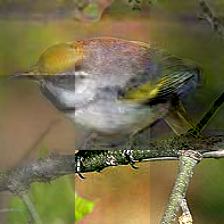}&
			\includegraphics[width=\imgwidthb \linewidth ]
			{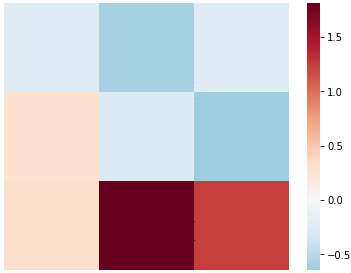}
			\\
			
			\includegraphics[width=\imgwidtha \linewidth ]
			{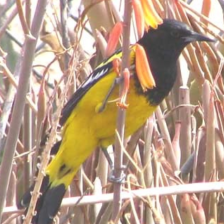}& 
			\includegraphics[width=\imgwidtha \linewidth ]
			{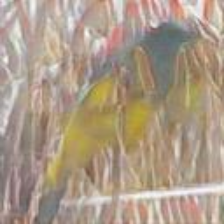}&
			\includegraphics[width=\imgwidtha \linewidth ]
			{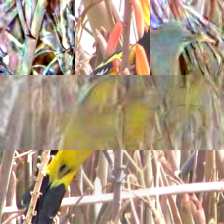}&
			\includegraphics[width=\imgwidthb \linewidth ]
			{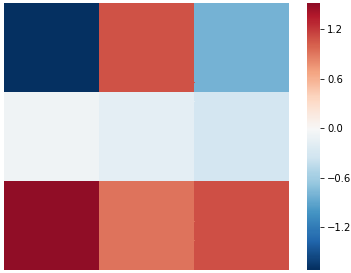}
			\\
			
			\includegraphics[width=\imgwidtha \linewidth ]
			{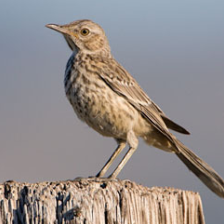}&
			\includegraphics[width=\imgwidtha \linewidth ]
			{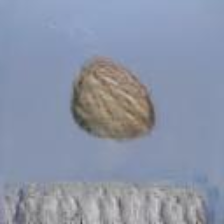}&
			\includegraphics[width=\imgwidtha \linewidth ]
			{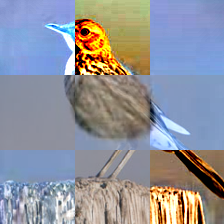}&
			\includegraphics[width=\imgwidthb \linewidth ]
			{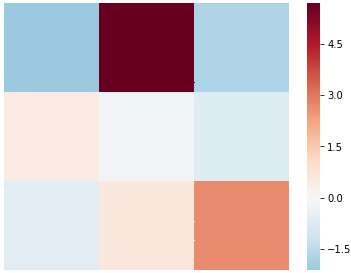}
			\\
			
			\includegraphics[width=\imgwidtha \linewidth ]
			{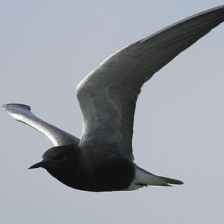}& 
			\includegraphics[width=\imgwidtha \linewidth ]
			{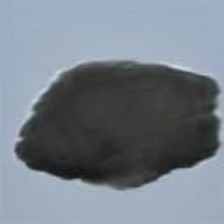}&
			\includegraphics[width=\imgwidtha \linewidth ]
			{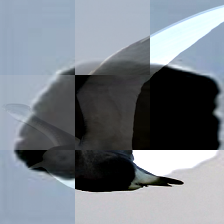}&
			\includegraphics[width=\imgwidthb \linewidth ]
			{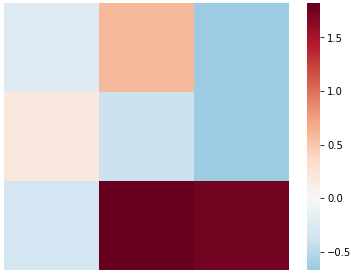}
			\\
		\end{tabular} 
		\caption{
			Samples of original image, generated image, fused image,  and mixing weight $\mathbf{w}$. Higher weight (red) means more of the original image was used, and lower weight (blue) means more generated image. %
		}\label{fig:result-samples}
	\end{figure}

	\begin{figure}[htb!]
		\centering
		\begin{minipage}{0.48\columnwidth}
			\centering
			\includegraphics[clip, width=\textwidth] {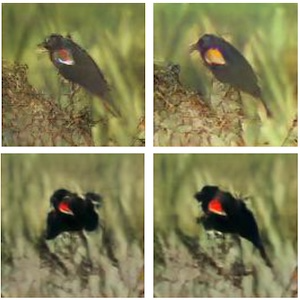}
			\caption{By fluctuating the input noise, FinetuneGAN can obtain slightly different variants of the original image, which we show in Figure~\ref{fig:fintune-gan-samples}(a). }\label{fig:other-noise}
		\end{minipage}\hfill
		\begin{minipage}{0.48\columnwidth}
			\centering
			\includegraphics[clip, width=\textwidth ]{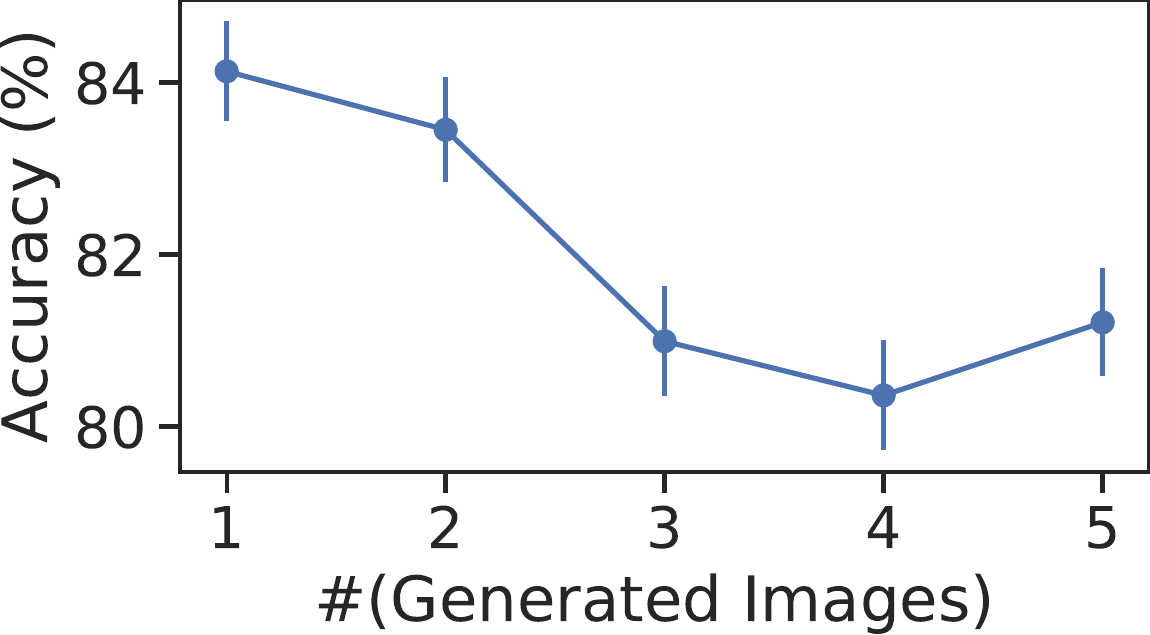} 
			\caption{
				MetaIRNet can in theory increase the number of fused images by using  images generated by FinetuneGAN with different random noise values. However, this does not increase accuracy, 
				presumably because they are conditioned on the same image so adding many of them does not increase diversity.
			}\label{fig:naug}
		\end{minipage}
	\end{figure}

	\xhdr{Increasing the number of generated examples.}  
	Does our method benefit by increasing the number of examples generated by FinetuneGAN with different random noise values (Figure~\ref{fig:other-noise})? We tried $n_{aug}=1, 2, 3, 4, 5$ on CUB and the accuracies are shown in Figure~\ref{fig:naug}. Having too many augmented images seems to bias the classifier, and we conclude that the performance gain is marginal or even harmful when increasing $n_{aug}$. This effect could be because all generated images are conditioned on the same original image, so adding many of them does not significantly increase diversity.
    
\xhdr{Comparing with other meta-learning classifiers.}
It is a convention in our community to compare any new technique with previous methods using the accuracies reported in the corresponding literature. The accuracies in Table~\ref{tbl:results-main}, however, cannot be directly compared with other papers' reported accuracies as we use ImageNet-pre-trained CNNs. While it is a natural design decision for us to use the pretrained model because our focus is how to use ImageNet pre-trained generators for improving fine-grained one-shot classification, which assumes ImageNet as an available resource off-the-shelf,  much of the one-shot learning literature focuses on improving the one-shot algorithms themselves and thus trains from scratch. To provide a comparison, we cite a benchmark study~\cite{chen2019closer} reporting accuracy of other well-known meta-learners~\cite{vinyals2016matching,finn2017model,sung2018learning} on the CUB dataset. To compare with these scores, we trained our MetaIRNet and the ProtoNet baseline using the same four-layered CNN. Our MetaIRNet achieved an accuracy  of $65.86\%\pm0.72$, which is higher than  ProtoNet~($63.50\%\pm0.70$), MatchingNet~($61.16\%\pm0.89$\cite{chen2019closer}), MAML~($55.92\%\pm0.95$\cite{chen2019closer}), and RelationNet~($62.45\%\pm0.98$\cite{chen2019closer}). We note that this comparison is not totally fair because we use images generated from a generator pre-trained from ImageNet, so one can argue that we use more data than others. However, our contribution is not to establish a new state-of-the-art score but to present the idea of transferring an ImageNet pre-trained GAN for improving one shot classifiers, so we believe this is still informative as it provides a reference score for future work to compare to.
	
	\xhdr{Results on NAB.}
	We also 
	performed similar experiments on the NAB dataset, which is more than four times
	larger than CUB, and the results are shown in the last column of
	Table~\ref{tbl:results-main}. We observe  similar results as on CUB, and that our method improves  classification accuracy
	from a ProtoNet baseline (89.19\% vs. 87.91\%).

\begin{figure}[t!]
	\centering
	\includegraphics[clip, width=\columnwidth ]{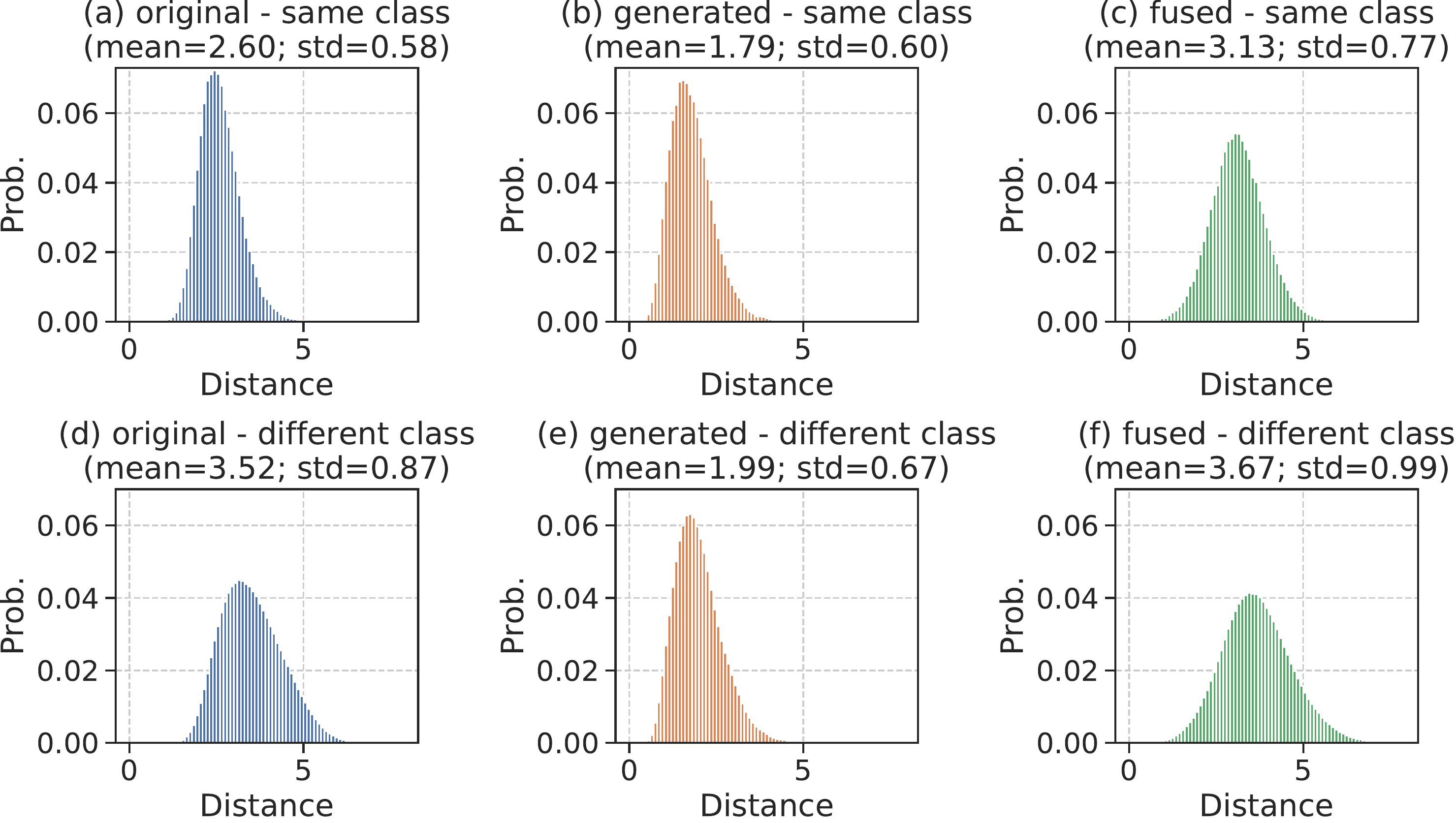} 
	\caption{
		Distribution of pairwise distances divided by intra-class (same-class) pairwise comparisons and inter-class (different-class) pairwise comparisons.  Similar to Figure~\ref{fig:analysis-dist}, our fused images have a wider pairwise distribution while generated images are not as diverse as originals, but the amount of increase from the original is more on the same-class comparisons.  
	}\label{fig:analysis-dist-class}
\end{figure}

	\section{Analysis}\label{sec:analysis}
	Our proposed technique for reinforcing images generated from fine-tuned GANs improved the few-shot recognition accuracy, but what causes this performance improvement? We  hypothesized that the images generated by GANs are not diverse enough on their own, and our learned technique that mixes them with  original images helps to diversify the dataset. To validate this hypothesis, we perform several studies investigating the diversity of three image sets: original images, images generated by GANs, and images fused by our method. To measure diversity, we use pairwise distance distributions and principal component analysis (PCA). 
	
	\subsection{Pairwise distance distribution}\label{sec-pairwise-dist} 
	One way of quantifying the diversity of an image set is to compute the distance between all possible pairs of images, and then examine the resulting distribution. We compute the Euclidean distances of all possible pairs of images in a set using pretrained CNN representations.  If the distribution of the pairwise distances of a set is longer-tailed than others, then we regard the set as more diverse. Figures~\ref{fig:analysis-dist}(a), (b), and (c) plot the distributions of original, generated, and fused images, respectively, using the CUB dataset. We observe that generated images (Figure \ref{fig:analysis-dist}(b)) do not increase pairwise distances from the original set (Figure \ref{fig:analysis-dist}(a)), and actually lower the mean distance from 3.50 to 1.99 and standard deviation from 0.87 to 0.67. In contrast, our fused images (Figure~\ref{fig:analysis-dist}(c)) slightly diversify the original set (Figure~\ref{fig:analysis-dist}(a)), and increase the mean distance from 3.50 to 3.66 and the standard deviation from 0.87 to 0.99.
	
	\subsection{Eigenvalues of PCA}
	We also employ a quantitative measure of
	the diversity of the data. 
	For a given image set, we compute its covariance matrix
	and apply principal component analysis (PCA). 
	PCA can help interpret a high dimensional space by decomposing it into orthogonal subspaces based on the variance of the data. The largest eigenvalues generated by PCA
	are a measure of the variance in the original
	dataset.
		We show a plot of the largest eigenvalues, sorted in decreasing order, 
		in Figure~\ref{fig:result-eigenvalue}. The figure confirms that the generated images have significantly lower eigenvalues than the original and fused images, and the fused images have slightly higher eigenvalues than the original (except for the highest eigenvalue). These observations indicate that the generated images are not diverse on their own, but our technique of fusing makes them at least as diverse as the originals. 
	
	        \subsection{Inter- and Intra-Class Diversity}
                Figure~\ref{fig:analysis-dist-class} shows the same histograms of pairwise distances, but
                split into image pairs that are within the same class and pairs that are in different classes.
                For
                the same type of images, it is intuitive that
                same-class distances are lower than different-class
                distances. Moreover, for both same-class and
                different-class, the overall trend is the same as
                Sec.~\ref{sec-pairwise-dist}: generated images are not
                as diverse as the originals, but our fused images are
                more diverse. However, when we compare the
                distribution of the original and fused images, the
                same-class comparison sees a greater increase in distance
                than the different-class comparison:
                the mean of different-class distances increases from
                3.52 (Figure~\ref{fig:analysis-dist-class}(d)) to 3.68
                (Figure~\ref{fig:analysis-dist-class}(f)), for a 
                change of +0.16, while the mean of same-class
                distances increases from 2.60
                (Figure~\ref{fig:analysis-dist-class}(a)) to 3.12
                (Figure~\ref{fig:analysis-dist-class}(c)), for a 
                change of +0.53.
                This suggests  that our fused images
                significantly widen the manifolds per class by
                efficiently mixing the original and generated images.

	\section{Conclusion}\label{sec:conclusion}
	We introduce an effective way to employ an ImageNet-pre-trained image
	generator for the purpose of improving fine-grained one-shot
	classification when data is scarce. Our pilot study found that adjusting only scale
	and shift parameters in batch normalization can produce  visually
	realistic images. This technique works with a single image, making the method
	less dependent on the number of available images. Furthermore,
	although naively adding the generated images into the training set
	does not improve  performance, we show that it can improve 
	performance if we  mix generated with original
	images to create hybrid training exemplars. In order to learn the parameters of this mixing, we adapt a
	meta-learning framework.
	We implement this idea and demonstrate a consistent and significant
	improvement over several classifiers on two fine-grained benchmark
	datasets. Furthermore, our analysis suggests that the increase in performance may be because the mixed images  are more diverse than the original and the generated images.

    \section*{Acknowledgments}
	We thank Minjun Li and
	Atsuhiro Noguchi for helpful discussions. Part of this work was done while
	Satoshi Tsutsui was an intern at Fudan University. Yanwei Fu was supported in
	part by the NSFC project (62076067), and Science and Technology
	Commission of Shanghai Municipality Project (19511120700). 
	DC was supported in part by 
	the National Science Foundation (CAREER IIS-1253549),
	and the IU Emerging Areas of Research Project ``Learning: Brains,
	Machines, and Children.'' Yanwei Fu is the corresponding author.

	\appendices
	\ifCLASSOPTIONcaptionsoff
	\newpage
	\fi

	\bibliographystyle{IEEEtran}
	\bibliography{references}

\begin{IEEEbiography}[{\includegraphics[width=1in,height=1.25in,trim=5cm 5cm 5cm 5cm,clip,keepaspectratio]{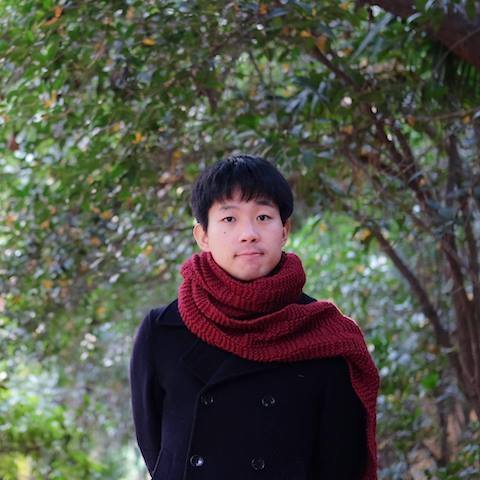}}]{Satoshi Tsutsui} received the BE degree from
Keio University, Tokyo, Japan, the MS degree
from Indiana University, Bloomington, Indiana,
and the PhD degree from Indiana University, in
2021, advised by Prof. David Crandall and Prof.
Chen Yu. He is currently a postdoctoral research
fellow with the National University of Singapore.
He is interested in computer vision for visual data
captured from wearable cameras (egocentric
vision).
\end{IEEEbiography}

    \begin{IEEEbiography}[{\includegraphics[width=1in,height=1.25in,clip,keepaspectratio]{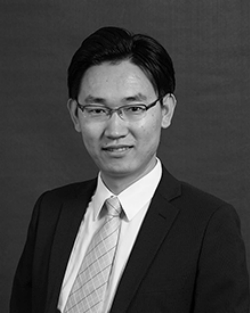}}]{Yanwei Fu} received his PhD degree from the Queen Mary University of London, in 2014. He worked as  post-doctoral research at Disney Research, Pittsburgh, PA, from 2015 to 2016. He is currently a tenure-track professor with Fudan University.  
He was appointed as the Professor of Special Appointment (Eastern Scholar) at Shanghai Institutions of Higher Learning.
He published more than 100 journal/conference papers including IEEE TPAMI, TMM, ECCV, and CVPR. His research interests are one-shot/meta learning,  and  learning based 3D reconstruction.
\end{IEEEbiography}

\begin{IEEEbiography}[{\includegraphics[width=1in,height=1.25in,clip,keepaspectratio]{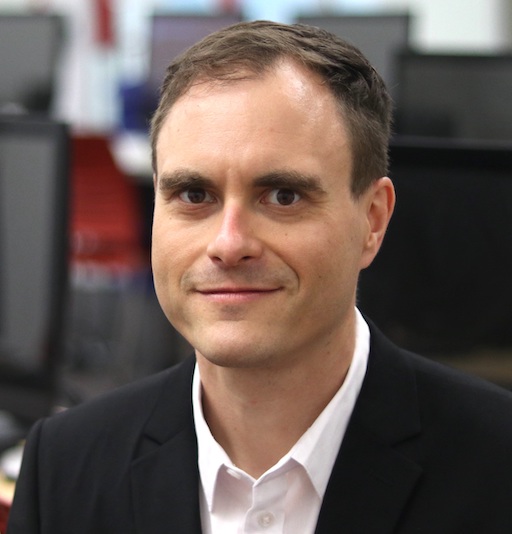}}]
  {\textbf{David Crandall}} is a Luddy Professor of Computer Science
  at Indiana University. He
  received the M.S. and Ph.D. degrees in Computer Science from Cornell
  University, Ithaca, NY in 2007 and 2008, respectively, and the B.S. and
  M.S. degrees in Computer Science and Engineering from the Pennsylvania
  State University, University Park, PA in 2001. His research interests include computer
  vision, machine learning, and data mining. He is the recipient of a
  National Science Foundation CAREER Award, two Google Faculty Research
  Awards, an IU Trustees Teaching Award, a Grant Thornton Fellowship, a Luddy named professorship,
  and numerous best paper awards and nominations. Currently he is an
  Associate Editor of IEEE TPAMI and IEEE TMM.
    \end{IEEEbiography}

\clearpage
\section*{Supplementary Material}
\subsection*{Five-shot Experiments}
Although our paper focuses on one-shot learning, we also try five-shot scenario.  We use the ImageNet pretrained ResNet18\cite{he2016deep} as a backbone. We also try the four layer CNN (Conv-4) without ImageNet pretraining  in order to compare with other reported scores in a benchmark study~\cite{chen2019closer}. The results are summarized in Table \ref{tbl:results-cub-5shot}. When we use ImageNet pretrained ResNet, our method is slightly (92.66\% v.s 92.83\%) better than the ProtoNet. Given that non meta-learning linear classifiers (softmax regression and logistic regression) can achieve more than 92\% accuracy, we believe that ImageNet features are already strong enough when used with five examples per class.

\begin{table}[tb!]
	\caption{5-way-5-shot Accuracy (\%) on CUB dataset.}
	\label{tbl:results-cub-5shot}
	\centering
	\begin{tabular}{lllc}
		\toprule
	     Method           & Base Network    & Initialization & Accuracy   \\
	     \midrule
		Nearest neighbor     & ResNet18 & ImageNet & $89.44 \pm 0.36$        \\ 
		Softmax regression  & ResNet18 & ImageNet & $92.28 \pm 0.30$        \\ 
		Logistic regression   & ResNet18 & ImageNet & $ {92.34 \pm 0.30}$        \\ 

         ProtoNet~\cite{snell2017prototypical} & ResNet18 & ImageNet & $92.97 \pm 0.31$        \\ 
         MetaIRNet (Ours) & ResNet18 & ImageNet  & $93.09 \pm 0.30$       \\
         \midrule
 		 MAML\cite{finn2017model}            & Conv-4   & Random     & $72.09 \pm 0.76$       \\ 
 		 MatchingNet~\cite{vinyals2016matching} & Conv-4   & Random  & $72.86 \pm 0.76$  \\ 
 		 RelationNet\cite{sung2018learning}     & Conv-4   & Random  & $76.11 \pm 0.69$\\ 
 		 ProtoNet~\cite{snell2017prototypical}  & Conv-4   & Random  & $80.75 \pm 0.46$         \\ 
 		 MetaIRNet (Ours)     & Conv-4   & Random  & $81.16 \pm 0.47$         \\ 
		\bottomrule
	\end{tabular}
\end{table}

\subsection*{An Implementation Detail: Class label input of BigGAN} Part of the noise $z$ used in BigGAN is class conditional, and we did not explcitly discuss this part in the main paper, so here we provide deatils.   We optimize the class
conditional embedding and regard it as part of the input noise.
Generally speaking, a conditional GAN uses input noise conditioned on
the label of the image to generate. BigGAN also follows this approach,
but our fine-tuning technique uses a single image to train. In other
words, we only have a single class label and can then optimize the
class embedding as part of the input noise. 
 \vspace{0pt}

\subsection*{More Experiments}

\subsubsection*{Image deformation baseline.} Image deformation net~\cite{img_deform_2019} also uses similar $3\times3$ patch based data augmentation learning. The key difference is that while that method augment
support image by fusing with external real images called a gallery set,
our model fuses with images synthesized by GANs. Further, to adapt a generic
pretrained GAN to a new domain, we introduce a technique of optimizing
only the noise z and BatchNorm parameters rather than the full generator, which
is not explored by deformation net~\cite{img_deform_2019}. We try this baseline by using a gallery set of random  images sampled from meta-traning set, and obtain 1-shot-5-way accuracies of $82.84 \pm 0.62$
on CUB and $88.42 \pm 0.59$ on NAB, which is higher than the
baselines but not as high as ours. 

\subsubsection*{$5\times5$ mixing weights.} We used $3\times3$ weights to mix the images by getting inspirations from previous work that pretrains CNNs by solving $3\times3$ Jigsaw puzzle~\cite{noroozi2016unsupervised} and from previous work~\cite{img_deform_2019} that mixes images with $3\times3$ patterns. While previous work suggests $3\times3$ works good in practice, there is no reason why we must use $3\times3$ instead of generic $i$ $\times$ $j$ where $i$ and $j$ are non-negative integers up to the height/width of the image. Finding the optimal $i$ and $j$ is not our interest, and we believe the answer ultimately depends on the dataset. Nonetheless, we experimented the $5\times5$ weights on CUB dataset, and got  1-shot-5-way accuracies of $83.63 \pm 0.61$, which is lower than that of $3\times3$ weights. 

\subsubsection*{Training from scratch or end-to-end.} It is an interesting direction to train the generator end-to-end and without ImageNet pretraining.  Theoretically, we can do end-to-end training of all components, but in practice we are limited by our GPU memory,  which is not large enough to hold both our model and BigGAN. In order to simulate the end-to-end and scratch training, we introduce two constrains 1) We simplified BigGAN with  one-quarter the number of channels and train from scratch so that we train the generator with relatively small meta-training set. 2) We do not propagate the gradient from classifier to the generator so that we do not have to put both models onto GPU. We apply our approach with a backbone of four-layer CNN with random initialization and achieved an 1-shot-5-way accuracy of  $63.77 \pm 0.71$ on CUB. 

\subsubsection*{Experiment on Mini-ImageNet.} Although our method is designed for fine-grained recognition, it is interesting to apply this to course-grained recognition. Because the public BigGAN model was trained on images including the meta-testing set of ImageNet, we cannot use it as it is.  Hence we train the simplified generator (see above paragraph) from scratch using meta-training set only. Using the backbone of ResNet18, the 1-shot-5-way accuracy on Mini-ImageNet is $53.97 \pm 0.63$ and $55.01 \pm 0.62$ for ProtoNet and MetaIRNet, respectively.

\end{document}

%% file: table1.tex
	\begin{table}
		\caption{5-way-1-shot accuracy (\%) on CUB and NAB datasets with ImageNet pre-trained ResNet18.}
		\label{tbl:results-main}
		\centering
                {\scriptsize{
		\begin{tabular}{@{}l cc@{}}
			\toprule
			Method (+Data Augmentation)   & CUB Acc. & NAB Acc. \\
			\midrule
			Nearest Neighbor    & $79.00\pm0.62$ &$80.58\pm0.59$\\
			Logistic Regression & $81.17\pm0.60$ &$82.70\pm0.57$\\
			Softmax Regression   & $80.77\pm0.60$ &$82.38\pm0.57$\\
			\midrule
			ProtoNet       & $81.73\pm0.63$  & $87.91\pm0.52$ \\ 
			ProtoNet  (+Flip)   & $82.66\pm0.61$  & $88.55\pm0.50$ \\ 
			ProtoNet  (+FinetuneGAN) & $79.40\pm0.69$  & $85.40\pm0.59$ \\ 
			ProtoNet  (+Gaussian) &$81.75\pm0.63$ & $87.90\pm0.52$\\ 
			ProtoNet  (+Mixup) & $82.65\pm0.59$ & $88.12\pm0.52$\\ 
			ProtoNet  (+Manifold Mixup) & $81.78\pm0.58$ & $85.31\pm0.51$\\ 
			ProtoNet  (+CutMix) &$80.81\pm0.71$ & $86.12\pm0.53$\\ 
			\midrule
			MetaIRNet (+FreezeDGAN) & $81.93\pm0.62$ & $88.26\pm0.57$ \\
			MetaIRNet (+Jitter) & $82.69\pm0.63$ &  $88.31\pm0.55$\\
			\midrule
			Ours: MetaIRNet (+FinetuneGAN)  & $84.13\pm0.58$ & $89.19\pm0.51$ \\
			Ours: MetaIRNet (+FinetuneGAN, Flip) & $\mathbf{84.80\pm0.56}$ & $\mathbf{89.57\pm0.49}$ \\
			\bottomrule
		\end{tabular}
                }}
	\end{table}

%% file: main.bbl
\begin{thebibliography}{10}
\providecommand{\url}[1]{#1}
\csname url@samestyle\endcsname
\providecommand{\newblock}{\relax}
\providecommand{\bibinfo}[2]{#2}
\providecommand{\BIBentrySTDinterwordspacing}{\spaceskip=0pt\relax}
\providecommand{\BIBentryALTinterwordstretchfactor}{4}
\providecommand{\BIBentryALTinterwordspacing}{\spaceskip=\fontdimen2\font plus
\BIBentryALTinterwordstretchfactor\fontdimen3\font minus
  \fontdimen4\font\relax}
\providecommand{\BIBforeignlanguage}[2]{{%
\expandafter\ifx\csname l@#1\endcsname\relax
\typeout{** WARNING: IEEEtran.bst: No hyphenation pattern has been}%
\typeout{** loaded for the language `#1'. Using the pattern for}%
\typeout{** the default language instead.}%
\else
\language=\csname l@#1\endcsname
\fi
#2}}
\providecommand{\BIBdecl}{\relax}
\BIBdecl

\bibitem{yuxiong2016eccv}
Y.~Wang and M.~Hebert, ``Learning from small sample sets by combining
  unsupervised meta-training with {CNNs},'' in \emph{{NeurIPS}}, 2016.

\bibitem{santoro2016meta}
A.~Santoro, S.~Bartunov, M.~Botvinick, D.~Wierstra, and T.~Lillicrap,
  ``Meta-learning with memory-augmented neural networks,'' in \emph{ICML},
  2016.

\bibitem{finn2017model}
C.~Finn, P.~Abbeel, and S.~Levine, ``Model-agnostic meta-learning for fast
  adaptation of deep networks,'' in \emph{{ICML}}, 2017.

\bibitem{img_deform_2019}
Z.~Chen, Y.~Fu, Y.-X. Wang, L.~Ma, W.~Liu, and M.~Hebert, ``Image deformation
  meta-networks for one-shot learning,'' in \emph{CVPR}, 2019.

\bibitem{goodfellow2014generative}
I.~Goodfellow, J.~Pouget-Abadie, M.~Mirza, B.~Xu, D.~Warde-Farley, S.~Ozair,
  A.~Courville, and Y.~Bengio, ``Generative adversarial nets,'' in
  \emph{{NeurIPS}}, 2014.

\bibitem{biggan}
A.~Brock, J.~Donahue, and K.~Simonyan, ``Large scale {GAN} training for high
  fidelity natural image synthesis,'' in \emph{ICLR}, 2019.

\bibitem{shmelkov2018good}
K.~Shmelkov, C.~Schmid, and K.~Alahari, ``How good is my {GAN}?'' in
  \emph{{ECCV}}, 2018.

\bibitem{metasatoshi19}
D.~C. Satoshi~Tsutsui, Yanwei~Fu, ``{Meta-Reinforced Synthetic Data for
  One-Shot Fine-Grained Visual Recognition},'' in \emph{{NeurIPS}}, 2019.

\bibitem{gulrajani2017improved}
I.~Gulrajani, F.~Ahmed, M.~Arjovsky, V.~Dumoulin, and A.~Courville, ``Improved
  training of {Wasserstein GANs},'' in \emph{{NeurIPS}}, 2017.

\bibitem{radford2015unsupervised}
A.~Radford, L.~Metz, and S.~Chintala, ``Unsupervised representation learning
  with deep convolutional generative adversarial networks,'' in \emph{ICLR},
  2016.

\bibitem{arjovsky2017wasserstein}
M.~Arjovsky, S.~Chintala, and L.~Bottou, ``Wasserstein {GAN},'' \emph{arXiv
  preprint arXiv:1701.07875}, 2017.

\bibitem{miyato2018spectral}
T.~Miyato, T.~Kataoka, M.~Koyama, and Y.~Yoshida, ``Spectral normalization for
  generative adversarial networks,'' in \emph{{ICLR}}, 2018.

\bibitem{wang2020stabilizing}
D.~Wang, X.~Qin, F.~Song, and L.~Cheng, ``Stabilizing training of generative
  adversarial nets via langevin stein variational gradient descent,''
  \emph{IEEE Trans Neural Netw Learn Syst}, 2020.

\bibitem{ojha2021few-shot-gan}
U.~Ojha, Y.~Li, C.~Lu, A.~A. Efros, Y.~J. Lee, E.~Shechtman, and R.~Zhang,
  ``Few-shot image generation via cross-domain correspondence,'' in
  \emph{{CVPR}}, 2021.

\bibitem{park2020swapping}
T.~Park, J.-Y. Zhu, O.~Wang, J.~Lu, E.~Shechtman, A.~A. Efros, and R.~Zhang,
  ``Swapping autoencoder for deep image manipulation,'' in \emph{{NeurIPS}},
  2020.

\bibitem{noguchi2019image}
A.~Noguchi and T.~Harada, ``Image generation from small datasets via batch
  statistics adaptation,'' in \emph{{ICCV}}, 2019.

\bibitem{wang2018transferring}
Y.~Wang, C.~Wu, L.~Herranz, J.~van~de Weijer, A.~Gonzalez-Garcia, and
  B.~Raducanu, ``Transferring {GANs}: generating images from limited data,'' in
  \emph{{ECCV}}, 2018.

\bibitem{de2017modulating}
H.~De~Vries, F.~Strub, J.~Mary, H.~Larochelle, O.~Pietquin, and A.~C.
  Courville, ``Modulating early visual processing by language,'' in
  \emph{{NeurIPS}}, 2017.

\bibitem{dumoulin2017learned}
V.~Dumoulin, J.~Shlens, and M.~Kudlur, ``A learned representation for artistic
  style,'' in \emph{{ICLR}}, 2017.

\bibitem{shrivastava2017learning}
A.~Shrivastava, T.~Pfister, O.~Tuzel, J.~Susskind, W.~Wang, and R.~Webb,
  ``Learning from simulated and unsupervised images through adversarial
  training,'' in \emph{{CVPR}}, 2017.

\bibitem{antoniou2018augmenting}
A.~Antoniou, A.~Storkey, and H.~Edwards, ``Augmenting image classifiers using
  data augmentation generative adversarial networks,'' in \emph{{ICANN}}, 2018.

\bibitem{zhang2018metagan}
R.~Zhang, T.~Che, Z.~Ghahramani, Y.~Bengio, and Y.~Song, ``{MetaGAN: An
  Adversarial Approach to Few-Shot Learning},'' in \emph{{NeurIPS}}, 2018.

\bibitem{gao2018low}
H.~Gao, Z.~Shou, A.~Zareian, H.~Zhang, and S.-F. Chang, ``Low-shot learning via
  covariance-preserving adversarial augmentation networks,'' in
  \emph{{NeurIPS}}, 2018.

\bibitem{chen2019closer}
W.-Y. Chen, Y.-C. Liu, Z.~Kira, Y.-C.~F. Wang, and J.-B. Huang, ``A closer look
  at few-shot classification,'' in \emph{{ICLR}}, 2019.

\bibitem{Triantafillou2020MetaDatasetAD}
E.~Triantafillou, T.~Zhu, V.~Dumoulin, P.~Lamblin, K.~Xu, R.~Goroshin,
  C.~Gelada, K.~Swersky, P.-A. Manzagol, and H.~Larochelle, ``Meta-dataset: A
  dataset of datasets for learning to learn from few examples,'' in
  \emph{ICLR}, 2020.

\bibitem{vinyals2016matching}
O.~Vinyals, C.~Blundell, T.~Lillicrap, D.~Wierstra \emph{et~al.}, ``Matching
  networks for one shot learning,'' in \emph{{NeurIPS}}, 2016.

\bibitem{snell2017prototypical}
J.~Snell, K.~Swersky, and R.~S. Zemel, ``Prototypical networks for few-shot
  learning,'' in \emph{{NeurIPS}}, 2017.

\bibitem{sung2018learning}
F.~Sung, Y.~Yang, L.~Zhang, T.~Xiang, P.~H. Torr, and T.~M. Hospedales,
  ``Learning to compare: Relation network for few-shot learning,'' in
  \emph{CVPR}, 2018.

\bibitem{bateni2020improved}
P.~Bateni, R.~Goyal, V.~Masrani, F.~Wood, and L.~Sigal, ``Improved few-shot
  visual classification,'' in \emph{{CVPR}}, 2020.

\bibitem{rusu2018meta}
A.~A. Rusu, D.~Rao, J.~Sygnowski, O.~Vinyals, R.~Pascanu, S.~Osindero, and
  R.~Hadsell, ``Meta-learning with latent embedding optimization,'' in
  \emph{ICLR}, 2018.

\bibitem{finn2018probabilistic}
C.~Finn, K.~Xu, and S.~Levine, ``Probabilistic model-agnostic meta-learning,''
  in \emph{NeurIPS}, 2018.

\bibitem{ravi2017optimization}
S.~Ravi and H.~Larochelle, ``Optimization as a model for few-shot learning,''
  in \emph{ICLR}, 2017.

\bibitem{wang2018low}
Y.-X. Wang, R.~Girshick, M.~Hebert, and B.~Hariharan, ``Low-shot learning from
  imaginary data,'' in \emph{{CVPR}}, 2018.

\bibitem{hariharan2017low}
B.~Hariharan and R.~Girshick, ``Low-shot visual recognition by shrinking and
  hallucinating features,'' in \emph{ICCV}, 2017.

\bibitem{chen2019imageblock_aaai}
Z.~Chen, Y.~Fu, K.~Chen, and Y.-G. Jiang, ``Image block augmentation for
  one-shot learning,'' in \emph{AAAI}, 2019.

\bibitem{schwartz2018delta}
E.~Schwartz, L.~Karlinsky, J.~Shtok, S.~Harary, M.~Marder, A.~Kumar, R.~Feris,
  R.~Giryes, and A.~Bronstein, ``Delta-encoder: an effective sample synthesis
  method for few-shot object recognition,'' in \emph{{NeurIPS}}, 2018.

\bibitem{krizhevsky2012imagenet}
A.~Krizhevsky, I.~Sutskever, and G.~Hinton, ``Imagenet classification with deep
  convolutional neural networks,'' in \emph{{NeurIPS}}, 2012.

\bibitem{zhong2020random}
Z.~Zhong, L.~Zheng, G.~Kang, S.~Li, and Y.~Yang, ``Random erasing data
  augmentation,'' in \emph{AAAI}, 2020.

\bibitem{cubuk2019autoaugment}
E.~D. Cubuk, B.~Zoph, D.~Mane, V.~Vasudevan, and Q.~V. Le, ``Autoaugment:
  Learning augmentation strategies from data,'' in \emph{{CVPR}}, 2019.

\bibitem{li2020differentiable}
Y.~Li, G.~Hu, Y.~Wang, T.~Hospedales, N.~M. Robertson, and Y.~Yang,
  ``Differentiable automatic data augmentation,'' in \emph{{ECCV}}, 2020.

\bibitem{gontijo-lopes2021tradeoffs}
R.~Gontijo-Lopes, S.~Smullin, E.~D. Cubuk, and E.~Dyer, ``Tradeoffs in data
  augmentation: An empirical study,'' in \emph{ICLR}, 2021.

\bibitem{dao2019kernel}
T.~Dao, A.~Gu, A.~Ratner, V.~Smith, C.~De~Sa, and C.~R{\'e}, ``A kernel theory
  of modern data augmentation,'' in \emph{ICML}, 2019.

\bibitem{zhang2018mixup}
H.~Zhang, M.~Cisse, Y.~N. Dauphin, and D.~Lopez-Paz, ``Mixup: Beyond empirical
  risk minimization,'' in \emph{ICLR}, 2018.

\bibitem{manifoldmixup19}
V.~Verma, A.~Lamb, C.~Beckham, A.~Najafi, I.~Mitliagkas, D.~Lopez-Paz, and
  Y.~Bengio, ``Manifold mixup: Better representations by interpolating hidden
  states,'' in \emph{ICML}, 2019.

\bibitem{yun2019cutmix}
S.~Yun, D.~Han, S.~J. Oh, S.~Chun, J.~Choe, and Y.~Yoo, ``Cutmix:
  Regularization strategy to train strong classifiers with localizable
  features,'' in \emph{ICCV}, 2019.

\bibitem{johnson2016perceptual}
J.~Johnson, A.~Alahi, and L.~Fei-Fei, ``Perceptual losses for real-time style
  transfer and super-resolution,'' in \emph{{ECCV}}, 2016.

\bibitem{WahCUB_200_2011}
C.~Wah, S.~Branson, P.~Welinder, P.~Perona, and S.~Belongie, ``{The
  Caltech-UCSD Birds-200-2011 Dataset},'' California Institute of Technology,
  Tech. Rep. CNS-TR-2011-001, 2011.

\bibitem{he2016deep}
K.~He, X.~Zhang, S.~Ren, and J.~Sun, ``Deep residual learning for image
  recognition,'' in \emph{CVPR}, 2016.

\bibitem{noroozi2016unsupervised}
M.~Noroozi and P.~Favaro, ``Unsupervised learning of visual representations by
  solving jigsaw puzzles,'' in \emph{{ECCV}}, 2016.

\bibitem{vgg}
K.~Simonyan and A.~Zisserman, ``Very deep convolutional networks for
  large-scale image recognition,'' in \emph{{ICLR}}, 2014.

\bibitem{van2015building}
G.~Van~Horn, S.~Branson, R.~Farrell, S.~Haber, J.~Barry, P.~Ipeirotis,
  P.~Perona, and S.~Belongie, ``Building a bird recognition app and large scale
  dataset with citizen scientists: The fine print in fine-grained dataset
  collection,'' in \emph{{CVPR}}, 2015.

\bibitem{kingma2014adam}
D.~P. Kingma and J.~Ba, ``Adam: A method for stochastic optimization,'' in
  \emph{ICLR}, 2015.

\bibitem{mo2020freeze}
S.~Mo, M.~Cho, and J.~Shin, ``Freeze the discriminator: a simple baseline for
  fine-tuning {GANs},'' in \emph{CVPR Workshop AICC}, 2020.

\end{thebibliography}
